\documentclass[11pt,a4paper]{article}
\usepackage[hyperref]{emnlp2018}
\usepackage{times}
\usepackage{latexsym}
\usepackage{amsmath,amsthm,amsfonts,amssymb,amsmath,enumitem}
\usepackage{bm}
\usepackage{url}
\usepackage{footnote}

\usepackage{graphicx}
\usepackage{booktabs}
\usepackage{multirow}
\usepackage{subfig}
\usepackage{tabularx}

\aclfinalcopy 
\def\aclpaperid{665} 

\title{Building Context-aware Clause Representations for Situation Entity Type Classification}

\author{Zeyu Dai, Ruihong Huang \\
	   Department of Computer Science and Engineering \\
       Texas A\&M University \\
  {\tt \{jzdaizeyu, huangrh\}@tamu.edu}}

\date{}

\begin{document}
\maketitle
\begin{abstract}
Capabilities to categorize a clause based on the type of situation entity (e.g., events, states and generic statements) the clause introduces to the discourse can benefit many NLP applications.
Observing that the situation entity type of a clause depends on discourse functions the clause plays in a paragraph and the interpretation of discourse functions depends heavily on paragraph-wide contexts, we propose to build context-aware clause representations for predicting situation entity types of clauses. 
Specifically, we propose a hierarchical recurrent neural network model to read a whole paragraph at a time and jointly learn representations for all the clauses in the paragraph by extensively modeling context influences and inter-dependencies of clauses.
Experimental results show that our model achieves the state-of-the-art performance for clause-level situation entity classification on the genre-rich MASC+Wiki corpus, which approaches human-level performance.
\end{abstract}

\section{Introduction} 
Clauses in a paragraph play different discourse and pragmatic roles and have different aspectual properties~\cite{smith1991,verkuyl2013compositional} accordingly.
We aim to categorize a clause based on its aspectual property and more specifically, based on the type of Situation Entity (SE)\footnote{The Situation Entity (SE) type of a clause is defined with respect to three situation-related features: the main NP referent type (specific or generic), fundamental aspectual class (stative or dynamic), and whether the situation evoked is episodic or habitual~\cite{friedrich2014situation}.} (e.g., events, states, generalizing statements and generic statements) the clause introduces to the discourse, following the recent work by~\cite{friedrich2016situation}.
Understanding SE types of clauses is beneficial for many NLP tasks, including discourse mode identification\footnote{E.g., EVENTs and STATEs are dominant in \textit{narratives} while GENERALIZINGs and GENERICs are dominant in \textit{informative} discourses.} \cite{smith2003,smith2005}, text summarization, information extraction and question answering. 


The situation entity type of a clause reflects discourse roles the clause plays in a paragraph and discourse role interpretation depends heavily on paragraph-wide contexts.
Recently, \citet{friedrich2016situation} used insightful syntactic-semantic features extracted from the target clause itself for SE type classification, which has achieved good performance across several genres when evaluated on the newly created large dataset MASC+Wiki.
In addition, \citet{friedrich2016situation} implemented a sequence labeling model with conditional random fields (CRF)~\cite{lafferty2001conditional} for fine-tuning a sequence of predicted SE types. 
However, other than leveraging common SE label patterns (e.g., GENERIC clauses tend to cluster together.), this approach largely ignored the wider contexts a clause appears in when predicting its SE type.

To further improve the performance and robustness of situation entity type classification, we argue that 
we should consider influences of wider contexts more extensively, not only by fine-tuning a sequence of SE type predictions,
but also in deriving clause representations and obtaining precise individual SE type predictions. 
For example, we distinguish GENERIC statements from GENERALIZING statements depending on if a clause expresses general information over classes or kinds instead of specific individuals. 
We recognize the latter two clauses in the following paragraph as GENERALIZING because both clauses describe situations related to the Amazon river: 

(1):\label{P1} {\it [Today, the Amazon river is experiencing a crisis of overfishing.]$_{\bf STATE}$ [Both subsistence fishers and their commercial rivals compete in netting large quantities of pacu,]$_{\bf GENERALIZING}$ [which bring good prices at markets in Brazil and abroad.]$_{\bf GENERALIZING}$}

If we ignore the wider context, the second clause can be wrongly recognized as GENERIC easily since ``fishers'' usually refer to one general class rather than specific individuals. 
However, considering the background introduced in first clause, ``fishers'' here actually refer to the fishers who fish on Amazon river which become specific individuals immediately.

Therefore, we aim to build context-aware clause representations dynamically which are informed by their paragraph-wide contexts. 
Specifically, we propose a hierarchical recurrent neural network model to read a whole paragraph at a time and jointly learn representations for all the clauses in the paragraph.
Our paragraph-level model derive clause representations by modeling inter-dependencies between clauses within a paragraph.
In order to further improve SE type classification performance, we also add an extra CRF layer at the top of our paragraph-level model to fine-tune a sequence of SE type predictions over clauses \cite{friedrich2016situation}, which however is not our contribution.

Experimental results show that our paragraph-level neural network model greatly improves the performance of SE type classification on the same MASC+Wiki~\cite{friedrich2016situation} corpus and achieves robust performance close to human level.
In addition, the CRF layer further improves the SE type classification results, but by a small margin. 
We hypothesize that situation entity type patterns across clauses may have been largely captured by allowing the preceding and following clauses to influence semantic representation building for a clause in the paragraph-level neural net model.

\section{Related Work}
\subsection {Linguistic Categories of SE Types}
The situation entity types annotated in the MASC+Wiki corpus \cite{friedrich2016situation} were initially introduced by~\citet{smith2003}, which were then extended by~\cite{palmer2007sequencing,friedrich2014situation}.
The situation entity types can be divided into the following broad categories:
\begin{itemize}
\vspace{-.05in}
\item \textbf{Eventualities} (EVENT, STATE and REPORT):  for clauses representing actual happenings and world states.
STATE and EVENT are two fundamental aspectual classes of a clause~\cite{siegel2000learning} which can be distinguished by the semantic property of dynamism.
REPORT is a subtype of EVENT for quoted speech. 
\vspace{-.1in}
\item \textbf{General Statives} (GENERIC and GENERALIZING): for clauses that express general information over classes or kinds, or regularities related to specific main referents. The type GENERIC is for utterances describing a general class or kind rather than any specific individuals (e.g., \textit{People love dogs.}). The type GENERALIZING is for habitual utterances that refer to ongoing actions or properties of specific individuals (e.g., \textit{Audubon educates the public.}). 
\vspace{-.1in}
\item \textbf{Speech Acts} (QUESTION and IMPERATIVE):  for clauses expressing two types of speech acts \cite{searle1969speech}.
\end{itemize}

\subsection{Situation Entity (SE) Type Classification}
Although situation entities have been well-studied in linguistics, there were only several previous works focusing on data-driven SE type classification using computational methods. 
\citet{palmer2007sequencing} first implemented a maximum entropy model for SE type classification relying on  words, POS tags and some linguistic cues as main features.
This work used a relatively small dataset (around 4300 clauses) and did not achieve satisfied performance (around 50\% of accuracy).

To bridge the gap, \citet{friedrich2016situation} created a much larger dataset MASC+Wiki (more than 40,000 clauses) and achieved better SE type classification performance (around 75\% accuracy) by using rich features extracted from the target clause.
The feature sets include POS tags, Brown cluster features, syntactic and semantic features of the main verb and main referent as well as features indicating the aspectual nature of a clause.
\citet{friedrich2016situation} further improved the performance by implementing a sequence labeling (CRF) model to fine-tune a sequence of SE type predictions and noted that much of the performance gain came from modeling the label pattern that 
GENERIC clauses often occur together. 
In contrast, we focus on deriving dynamic clause representations informed by paragraph-level contexts and model context influences more extensively. 

\citet{becker2017classifying} proposed a GRU based neural network model that predicts the SE type for one clause each time, by encoding the content of the target clause using a GRU and incorporating several sources of context information, including contents and labels of preceding clauses as well as genre information, using additional separate GRUs~\cite{GRU}.
This model is different from our approach that processes one paragraph (with a sequence of clauses) at a time and extensively models inter-dependencies of clauses.

Other related tasks include predicting aspectual classes of verbs~\cite{friedrich2014automatic}, classifying genericity of noun phrases~\cite{reiter2010identifying} and predicting clause habituality~\cite{friedrich2015automatic}.  

\begin{figure*}[h]
\includegraphics[height=105mm,width=\textwidth]{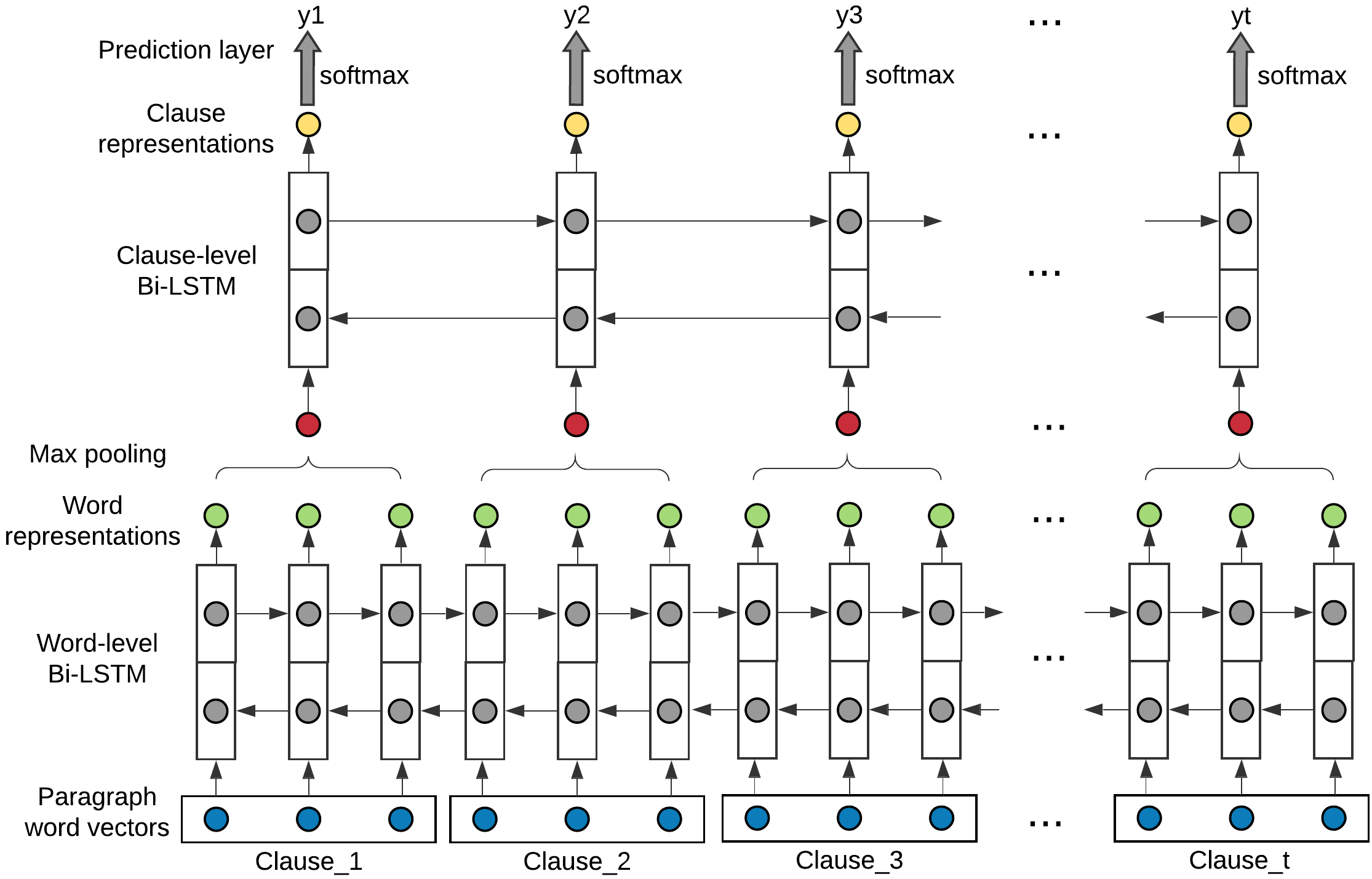}
\caption{The Paragraph-level Model Architecture for Situation Entity Type Classification.}
\label{paragraph_setting}
\end{figure*}

\subsection {Paragraph-level Sequence Labeling}
Learning latent representations and predicting a sequence of labels from a long sequence of sentences (clauses), such as a paragraph, is a challenging task.
Recently, various neural network models, including Convolution Neural Network (CNN)~\cite{wang2017sequence}, Recurrent Neural Network (RNN) based models~\cite{wang2015part,chiu2016named,huang2015bidirectional,ma2016end,lample2016neural} and Sequence to Sequence models~\cite{vaswani2016supertagging,zheng2017joint}, have been applied to the general task of sequence labeling.
Among them, the bidirectional LSTM (Bi-LSTM) model~\cite{schuster1997bidirectional} has been widely used to process a paragraph for applications such as language generation~\cite{li2015hierarchical}, dialogue systems~\cite{serban2016building} and text summarization~\cite{nallapati2016abstractive}, because of its capabilities in modeling long-distance dependencies between words. 
In this work, we use two levels of Bi-LSTMs connected by a max-pooling layer to abstract clause representations by extensively modeling paragraph-wide contexts and inter-dependencies between clauses.

\section{The Hierarchical Recurrent Neural Network for SE Type Classification}
We design an unified neural network to extensively model word-level dependencies
as well as clause-level dependencies in deriving clause representations for SE type prediction.
Figure \ref{paragraph_setting} shows the architecture of the proposed paragraph-level neural network model which includes two Bi-LSTM layers, one max-pooling layer in between and one final softmax prediction layer.

Given the word sequence of one paragraph as input, the word-level Bi-LSTM will firstly generate a sequence of hidden states as word representations, then a max-pooling layer will be applied to abstract clause embeddings from word representations within a clause. 
Next, another clause-level Bi-LSTM will run over the sequence of clause embeddings and derive final clause representations by further modeling semantic dependencies between clauses within a paragraph. 
The softmax prediction layer will then predict a sequence of situation entity (SE) types with one label for each clause, based on the final clause representations.

\vspace{.1in}
\noindent{\bf Word Vectors:} 
To transform the one-hot representation of each word into its distributed word vector~\cite{mikolov2013distributed}, we used the pre-trained 300-dimension Google English word2vec embeddings\footnote{Downloaded from \url{ https://docs.google.com/uc?id=0B7XkCwpI5KDYNlNUTTlSS21pQmM}}.
For the words which are not included in the vocabulary of Google word2vec, we randomly initialize their word vectors with each dimension sampled from the range $[-0.25, 0.25]$.

For situation entity type classification, it is important to recognize certain types of words such as punctuation marks (e.g., ``?'' for QUESTION and ``!'' for IMPERATIVE) as well as entities such as locations and time values.  
We therefore created feature-rich word vectors by concatenating word embeddings with parts-of-speech (POS) tag and named-entity (NE) tag one-hot embeddings\footnote{Our feature-rich word vectors are of dimension 343, including 300 dimensions for Google word2vec + 36 dimensions for POS tags + 7 dimensions for NE tags. We used the Stanford CoreNLP to generate POS tags and NE tags.}.

\vspace{.1in}
\noindent{\bf Deriving Clause Representations:}
In designing the model, we focus on building clause representations that sufficiently leverage cues from paragraph-wide contexts for SE type prediction, including both preceding and following clauses in a paragraph. 
To process long paragraphs which may contain a number of clauses, we utilize a two-level bottom-up abstraction approach and progressively obtain the compositional representation of each word (low-level) and then compute a compositional representation of each clause (high-level), with a max-pooling layer in between. 

At both word-level and clause-level, we choose the Bi-LSTM as our basic neural net component for representation learning, mainly considering its ability to capture long-distance dependencies between words (clauses) and to integrate influences of context words (clauses) from both directions. 

Given a word sequence $X = (x_1,x_2,...,x_L)$ in a paragraph as the input, the word-level Bi-LSTM will process the input paragraph by using two separate LSTMs, one processes the word sequence from the left to right while the other processes the sequence from the right to left. 
Therefore, at each word position $t$, we obtain two hidden states $\overrightarrow{h_t}, \overleftarrow{h_t}$ and concatenate them to get the word representation $h_t = [\overrightarrow{h_t}, \overleftarrow{h_t}]$.
Then we apply the max-pooling operation over the sequence of word representations for words within a clause in order to get the initial clause embedding:
\vspace{-.3in}
\begin{center}
\begin{align}
h_{Clause}[j] =\max_{t=Clause\_start}^{Clause\_end}h_t[j]\quad \\
where, 1 \leq j \leq hidden\_unit\_size
\end{align}
\end{center}

Next, the clause-level Bi-LSTM will process the sequence of initial clause embeddings in a paragraph and generate refined hidden states $\overrightarrow{h_{Clause\_t}}$ and $\overleftarrow{h_{Clause\_t}}$ at each clause position $t$. 
Then, we concatenate the two hidden states for a clause to get the final clause representation $h_{Clause\_t} = [\overrightarrow{h_{Clause\_t}}, \overleftarrow{h_{Clause\_t}}]$. 

\vspace{.1in}
\noindent{\bf Situation Entity Type Classification:}
Finally, the prediction layer will predict the situation entity type for each clause by applying the softmax function to its clause representation:
\vspace{-.2in}
\begin{center}
\begin{align}
y_{t} = softmax(W_y*h_{Clause\_t}+b_y) 
\end{align}
\end{center}
\vspace{-.1in}

\subsection{Fine-tune Situation Entity Predictions with a CRF Layer}
\begin{figure}[h]
\includegraphics[height=30mm,width=77mm]{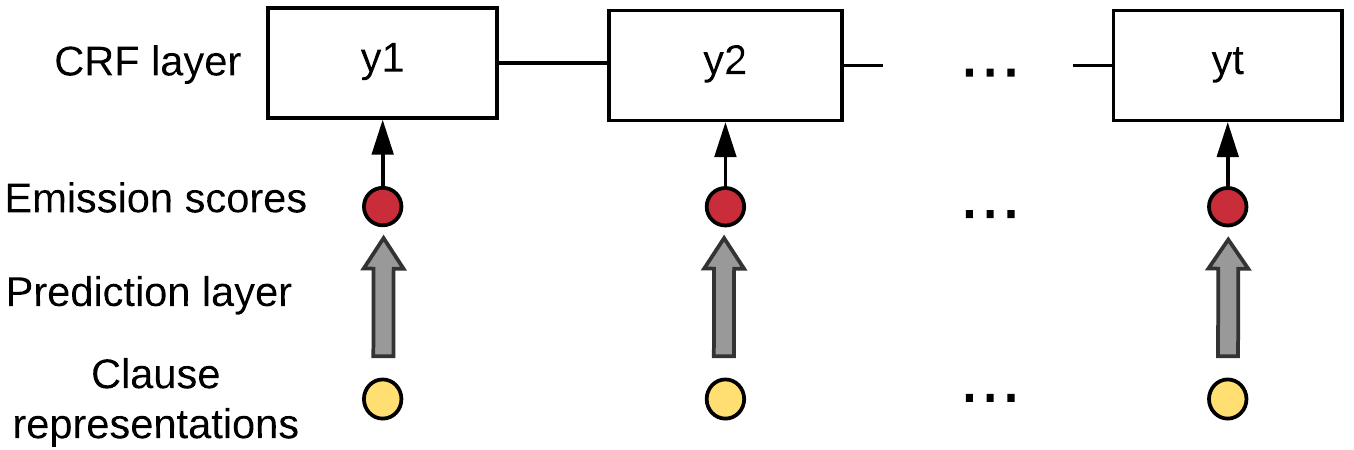}
\caption{Fine-tune a Situation Entity Label Sequence with a CRF layer.}
\label{CRF_layer}
\end{figure}

Previous studies~\cite{friedrich2016situation,becker2017classifying} show that there exist common SE label patterns between adjacent clauses.
For example, \citet{friedrich2016situation} reported the fact that GENERIC sentences usually occur together in a paragraph. 
Following \cite{friedrich2016situation}, in order to capture SE label patterns in our hierarchical recurrent neural network model, we add a CRF layer at the top of the softmax prediction layer (shown in figure \ref{CRF_layer}) to fine-tune predicted situation entity types. 

The CRF layer will update a state-transition matrix, which can effectively adjust the current label depending on its preceding and following labels. 
Both the training and decoding procedures of the CRF layer can be conducted efficiently using the Viterbi algorithm.
With the CRF layer, the model jointly assigns a sequence of SE labels, one label per clause, by considering individual clause representations as well as common SE label patterns. 

\subsection{Parameter Settings and Model Training}
We finalized hyperparameters based on the best performance with 10-fold cross-validation on the training set.
The word vectors were fixed during model training. 
Both word representations and clause representations in the model are of 300 dimensions, and all the Bi-LSTM layers contain 300 hidden units as well. 
To avoid overfitting, we applied dropout mechanism~\cite{hinton2012improving} with dropout rate of 0.5 to both input and output vectors of Bi-LSTM layers.
To deal with the exploding gradient problem in LSTMs training, we utilized gradient clipping~\cite{pascanu2013difficulty} with gradient L2-norm threshold of 5.0 and used L2 regularization with $\lambda=10^{-4}$ simultaneously. 
These parameters remained the same for all our proposed models including our own baseline models.

We chose the standard cross-entropy loss function for training our neural network models and adopted Adam~\cite{kingma2014adam} optimizer with the initial learning rate of 0.001 and the batch size\footnote{Counted as the number of SEs rather than paragraph instances.} of 128.
All our proposed models were implemented with Pytorch\footnote{\url{http://pytorch.org/}} and converged to the best result within 40 epochs.
Note that to diminish the effects of randomness in training neural network models and report stable experimental results, we ran each of the proposed models as well as our own baseline models ten times and reported the averaged performance across the ten runs.

\section{Evaluation}
\subsection{Dataset and Preprocessing}
\textbf{The MASC+Wiki Corpus}:  
We evaluated our neural network model on the MASC+Wiki corpus\footnote{\url{www.coli.uni-saarland.de/projects/sitent/page.php?id=resources}} \cite{friedrich2016situation}, which contains more than 40,000 clauses and is the largest annotated dataset for situation entity type classification. 
The MASC+Wiki dataset is composed of documents from Wikipedia and MASC~\cite{ide2008masc} covering as many as 13 written genres (e.g., news, essays, fiction, etc).
Table \ref{dataset statistics} shows statistics of the dataset, from which you can see that the SE type distribution is highly imbalanced.
The majority SE type of MASC documents is STATE while the majority SE type of Wikipedia documents is GENERIC.
To make our results comparable with previous works~\cite{friedrich2016situation,becker2017classifying}, we used the same 80:20 train-test split with balanced genre distributions.

\textbf{Preprocessing}: As described in \cite{friedrich2016situation}, texts were split into clauses using SPADE \cite{soricut2003sentence}. 
There are 4,784 paragraphs in total in the corpus; and on average, each paragraph contains 9.6 clauses.
In figure \ref{impact of paragraph length}, the horizontal axis shows the distribution of paragraphs based on the number of clauses in a paragraph. 
The annotations of clauses are stored in separate files from the text files. 
To recover the paragraph contexts for each clause, we matched its content with the corresponding raw document.

\begin{table}[]
\centering
\begin{tabular}{|l|rr|r|}
\hline
SE type      & \multicolumn{1}{c}{MASC} & \multicolumn{1}{c|}{Wiki} & \multicolumn{1}{c|}{Count} \\ \hline
STATE        & 49.8\%                   & 24.3\%                    & 18337                      \\
EVENT        & 24.3\%                   & 18.9\%                    & 9688                       \\
REPORT       & 4.8\%                    & 0.9\%                     & 1617                       \\
GENERIC      & 7.3\%                    & 49.7\%                    & 7582                       \\
GENERALIZING & 3.8\%                    & 2.5\%                     & 1466                       \\
QUESTION     & 3.3\%                    & 0.1\%                     & 1056                       \\
IMPERATIVE   & 3.2\%                    & 0.2\%                     & 1046                       \\ \hline
\end{tabular}
\caption{MASC+Wiki Dataset Statistics.}
\label{dataset statistics}
\end{table}

\begin{table*}[t]
\centering
\begin{tabular}{|l|cc|ccccccc|}
\hline
\multicolumn{1}{|c|}{Model}             & Macro & Acc   & STA & EVE & REP & GENI & GENA & QUE & IMP \\ \hline
Humans                                 & 78.6  & 79.6  & 82.8  & 80.5  & 81.5   & 75.1    & 45.8         & 90.7     & 93.6       \\
CRF \cite{friedrich2016situation}                       & 71.2  & 76.4  & 80.6  & 78.6  & 78.9   & 68.3    & 29.4         & 84.4     & 75.3       \\
 \hline
Clause-level Bi-LSTM & 74.4  & 78.3  & 82.6  & 81.3  & 84.9   & 66.2    & 36.1         & 88.5       & 80.9       \\ \hline
Paragraph-level Model                   & 77.6  & 81.2  & 84.3  & \textbf{82.1}  & 85.3   & 76.4    & 43.2         & \textbf{90.8}     & 81.2       \\
Paragraph-level Model+CRF             & \textbf{77.8}  & \textbf{81.3}  & 84.3  & 82.0  & \textbf{85.7}   & \textbf{77.0}    & \textbf{43.5}         & 90.4       & \textbf{81.5}       \\ \hline
\end{tabular}
\caption{Situation Entity Type Classification Results on the Training Set of MASC+Wiki with 10-Fold Cross-Validation. We report accuracy (Acc), macro-average F1-score (Macro) and class-wise F1 scores for STATE (STA), EVENT (EVE), REPORT (REP), GENERIC (GENI), GENERALIZING (GENA), QUESTION (QUE) and IMPERATIVE (IMP).}
\label{CV result}
\end{table*}

\subsection{Systems for Comparisons}
We compare the performance of our neural network model with two recent SE type classification models on the MASC+Wiki corpus as well as humans' performance (upper bound).

\begin{itemize}
\vspace{-.1in}
\item CRF \cite{friedrich2016situation}: 
a CRF model that relies heavily on features extracted from the target clause itself.
\vspace{-.1in}
\item GRU \cite{becker2017classifying}: a GRU based neural network model that incorporates context information by using separate GRU units and predicts the SE type for one clause each time.
\vspace{-.3in}
\item Humans \cite{friedrich2016situation}: one annotator's performance when using two other annotators' annotation as ``gold labels''. It has been reported that labeling SE types is a non-trivial task even for humans.
\vspace{-.1in}
\end{itemize}

In addition, we implemented a clause-level Bi-LSTM model as our own baseline, which takes a single clause as its input.
Since there is only one clause, the upper Bi-LSTM layer shown in Figure \ref{paragraph_setting} is meaningless and removed in the clause-level Bi-LSTM model.

\subsection{Experimental Results}
Following the previous work~\cite{friedrich2016situation} on the same task and dataset, we report accuracy and macro-average F1-score across SE types on the test set of MASC+Wiki.

\begin{table}[]
\centering
\begin{tabular}{|l|cc|}
\hline
\multicolumn{1}{|c|}{Model}     & Macro & \multicolumn{1}{l|}{Acc} \\ \hline
CRF \cite{friedrich2016situation}                             & 69.3               & 74.7                      \\
GRU \cite{becker2017classifying}                             & 68.0 & 71.1 \\ \hline
Clause-level Bi-LSTM & 73.5               & 76.7                      \\ \hline
Paragraph-level Model           & 77.0               & 80.0                      \\
Paragraph-level Model + CRF     & \textbf{77.4}               & \textbf{80.7}                      \\ \hline
\end{tabular}
\caption{Situation Entity Type Classification Results on the Test Set of MASC+Wiki. We report accuracy (Acc) and macro-average F1 (Macro).}
\label{test result}
\end{table}

The first section of Table \ref{test result} shows the results of the previous works.
The second section shows the result of our implemented clause-level Bi-LSTM baseline, which already outperforms the previous best model.
This result proves the effectiveness of the Bi-LSTM + max pooling approach in clause representation learning~\cite{Conneau2017}.
The third section reports the performance of the paragraph-level models that uses paragraph-wide contexts as input.
Compared with the baseline clause-level Bi-LSTM model, the basic paragraph-level model achieves 3.5\% and 3.3\% of performance gains in macro-average F1-score and accuracy respectively.
Building on top of the basic paragraph-level model, the CRF layer further improves the SE type prediction performance slightly by 0.4\% and 0.7\% in macro-average F1-score and accuracy respectively. 
Therefore, our full model with the CRF layer achieves the state-of-the-art performance on the MASC+Wiki corpus.

\begin{table*}[]
\centering
\begin{tabular}{|l|cc|ccccccc|}
\hline
\multicolumn{1}{|c|}{Model}     & Macro & Acc  & STA  & EVE  & REP  & GENI & GENA & QUE  & IMP  \\ \hline
CRF \cite{friedrich2016situation}                              & 66.6  & 71.8 & 78.2 & 77.0 & 76.8 & 44.8 & 27.4 & 81.8 & 70.8 \\ \hline
Clause-level Bi-LSTM & 69.3  & 73.3 & 79.5 & 78.7 & 82.8 & 47.6 & 31.9 & 86.9 & 77.7 \\ \hline
Paragraph-level Model           & 73.2  & 77.2 & 81.5 & 80.1 & 83.2 & 64.7 & 37.2 & 88.1 & \textbf{77.8} \\
Paragraph-level Model+CRF     & \textbf{73.5}  & \textbf{77.4} & 81.5 & \textbf{80.3} & \textbf{83.7} & \textbf{66.5} & \textbf{37.4} & \textbf{88.5} & 76.7 \\ \hline
\end{tabular}
\caption{Cross-genre Classification Results on the Training Set of MASC+Wiki. We report accuracy (Acc), macro-average F1-score (Macro) and class-wise F1 scores.}
\label{cross-genre result}
\end{table*}

\section{Analysis}
\subsection{10-Fold Cross-Validation}
We noticed that the previous work~\cite{friedrich2016situation} did not publish the class-wise performance of their model on the test set, instead, they reported the detailed performance on the training set using 10-fold cross-validation. 
For direct comparisons, we also report our 10-fold cross-validation results\footnote{The original folds split used by \citet{friedrich2016situation} is not available. So we manually split folds by ourselves with even genre distribution across folds.} on the training set of MASC+Wiki.

Table \ref{CV result} reports the cross-validation classification results.
Consistently, our clause-level baseline model already outperforms the previous best model.
By exploiting paragraph-wide contexts, the basic paragraph-level model obtains consistent performance improvements across all the classes compared with the baseline clause-level prediction model, especially for the classes GENERIC and GENERALIZING, where the improvements are significant. 
After using the CRF layer to fine-tune the predicted SE label sequence, 
slight performance improvements were observed on the four small classes.
Overall, the full paragraph-level neural network model achieves the best macro-average F1-score of 77.8\% in predicting SE types, which not only outperforms all previous approaches but also reaches human-like performance on some classes. 

\subsection{Impact of Genre}
Considering that MASC+Wiki is rich in written genres, we additionally conduct cross-genre classification experiments, where we use one genre of documents for testing and the other genres of documents for training.
The purpose of cross-genre experiment is to see whether the model can work robustly across genres.

Table \ref{cross-genre result} shows cross-genre experimental results of our neural network models on the training set of MASC+Wiki by treating each genre as one cross-validation fold.
As we expected, both the macro-average F1-score and class-wise F1 scores are lower compared with the results in Table \ref{CV result} where in-genre data were used for model training as well. 
But the performance drop on the paragraph-level models is little, which clearly outperform the previous system \cite{friedrich2016situation} and the baseline model by a large margin.
As shown in Table \ref{cross-genre result by genre}, benefited from modeling wider contexts and common SE label patterns, our full paragraph-level model improves performance across almost all the genres.
The high performance in the cross-genre setting demonstrates the robustness of our paragraph-level model across genres.

\begin{table}[]
\centering
\begin{tabular}{|l|cc|c|}
\hline
Genre     & \multicolumn{1}{l}{Baseline} & Full Model & Humans \\ \hline
blog      & 66.7                         & \textbf{70.3}       & 72.9   \\
email     & 71.1                         & \textbf{71.5}       & 67.0   \\
essays    & 61.2                         & \textbf{64.1}       & 64.6   \\
ficlets   & 67.9                         & \textbf{68.8}       & 81.7   \\
fiction   & 70.2                         & \textbf{72.1}       & 76.7   \\
gov-docs  & 68.6                         & \textbf{68.9}       & 72.6   \\
jokes     & 70.0                         & \textbf{75.0}       & 82.0   \\
journal   & \textbf{66.7}                & 66.4       & 63.7   \\
letters   & 68.6                         & \textbf{71.2}       & 68.0   \\
news      & 70.4                         & \textbf{72.7}       & 78.6   \\
technical & 55.7                         & \textbf{60.5}       & 54.7   \\
travel    & 51.3                         & \textbf{53.6}       & 48.9   \\
wiki      & 55.2                         & \textbf{60.6}       & 69.2   \\ \hline
\end{tabular}
\caption{Cross-genre Classification Results by Genre on the Training Set of MASC+Wiki.
Baseline: Clause-level Bi-LSTM; Full Model: Paragraph-level Model + CRF. We report macro-average F1-score for each genre.}
\label{cross-genre result by genre}
\end{table}

\begin{figure}[t]
\centering
\includegraphics[height=50mm,width=60mm]{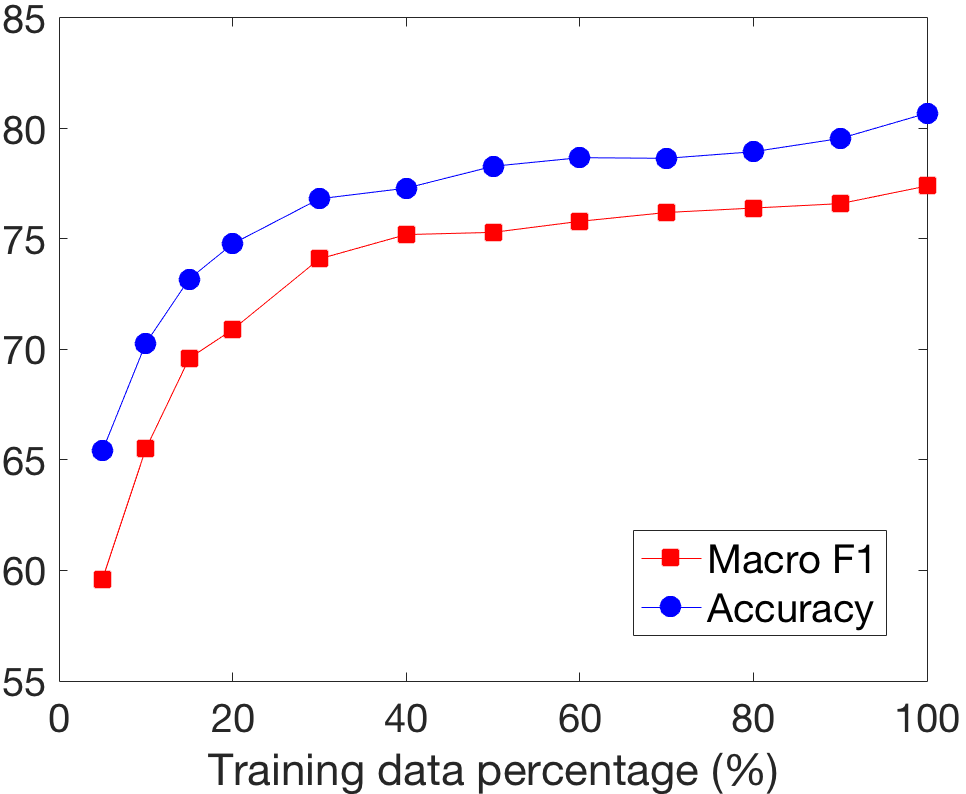}
\caption{Learning Curve of the Paragraph-level Model + CRF on MASC+Wiki.}
\label{learning curve}
\end{figure}

\begin{figure*}[h]
\includegraphics[height=66mm,width=\textwidth]{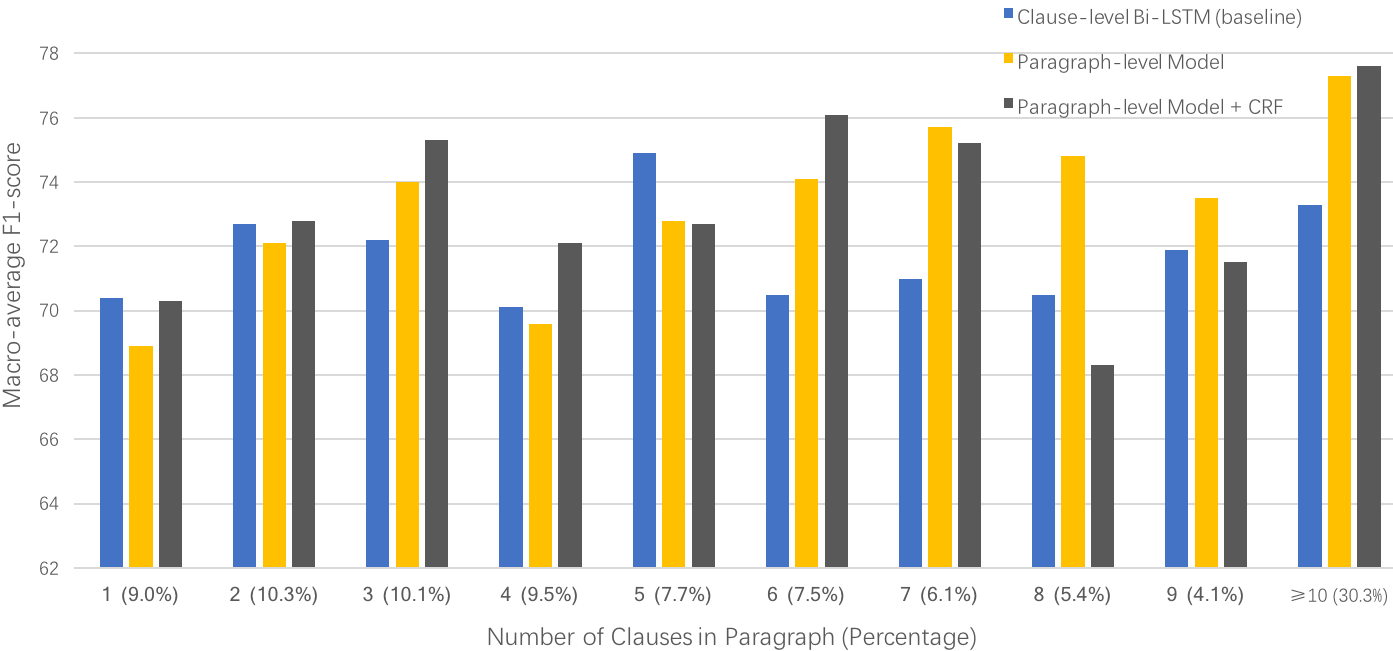}
\caption{Impact of Paragraph Lengths. We plot the macro-average F1-score for each paragraph length.}
\label{impact of paragraph length}
\end{figure*}

\subsection{Impact of Training Data Size}

In order to understand how much training data is required to train the paragraph-level model and obtain a good performance for SE type classification, we plot the learning curve shown in Figure \ref{learning curve} by training the full model several times using an increasing amount of training data.
The classification performance increased quickly before the amount of training data was increased to      
30\% of the full training set; then the learning curve starts to become saturated afterwards. 
We conclude that the paragraph-level model can achieve a high performance quickly without requiring a large amount of training data.

\subsection{Impact of Paragraph Length}

To study the influence of paragraph lengths to the performance of the paragraph-level models, we report the performance of our proposed models on subsets of the test set, with paragraphs divided based on the number of clauses in a paragraph. 
The histogram in Figure \ref{impact of paragraph length} compares performance of the two paragraph-level models and the baseline model. 
Note that the last bucket (paragraphs containing ten or more clauses) of the histogram is especially large and contains over 30\% of all the paragraphs in the test set.
Clearly, the paragraph-level model greatly outperforms the baseline clause-level model on paragraphs containing more than 6 clauses, which covers over 50\% of the test set. 
Adding the CRF layer further improves the performance of the paragraph-level model on long paragraphs (with 10 or more clauses),  while the influences to the performance are mixed on short paragraphs. 
Therefore, it is beneficial to model wider paragraph-level contexts and inter-dependencies between clauses for situation entity type classification, especially when processing long paragraphs.

\subsection{Impact of Discourse Connective Phrases}

As one aspect of modeling context influences and clause inter-dependencies in SE type identification, we investigated the role of discourse connective phrases in determining the SE type of clauses they connect. 
Our assumption is that discourse connectives are important to glue clauses together and removing them affects text coherence and information flow between clauses. 
Intuitively, the connective ``and'' may occur between two clauses with the same SE type; ``for example'' may indicate that the following clause is not GENERIC.
Therefore, we designed a pilot experiment to see whether discourse connective phrases are indispensable in building clause representations. 

In this pilot experiment, we extracted a list of 100 explicit discourse connectives. 
PDTB corpus \cite{Prasad08thepenn} and identified clauses that start with a discourse connecte. We found that 20.6\% of clauses in the MASC+Wiki corpus contain a discourse connective phrase.   
Then we ran the full paragraph-level model with one modification, i.e., disregarding words in connective phrases when conducting the max-pooling operation in equation (1), thus we did not consider discourse connective phrases directly when building a clause representation.

As shown in Table \ref{impact of connective phrases}, for clauses containing a discourse connective phrase, both macro-average F1-score and accuracy dropped due to the exclusion of discourse connective phrases.  
The performance was negatively influenced across all the SE types except the type of QUESTION and IMPERATIVE\footnote{A possible explanation is that recognizing QUESTION (IMPERATIVE) clauses mainly relies on seeing certain punctuation marks and key words, such as ``?'' (``!'') and ``why'' (``please''), which are independent from discourse connectives.}. 
The performance decreases on three SE types, REPORT, GENERIC and GENERALIZING, are noticeable. 
To some extent, this pilot study shows that modeling text coherence and the overall discourse structure of a paragraph is important in situation entity type classification.

\begin{table*}[]
\centering
\begin{tabular}{|cc|ccccccc|}
\hline
Macro & Acc  & STA  & EVE  & REP  & GENI & GENA & QUE  & IMP  \\ \hline
-1.2  & -0.9 & -1.0 & -0.8 & -2.3 & -3.4 & -2.2 & 0.5 & 0.3 \\ \hline
\end{tabular}
\caption{Impact of Discourse Connective Phrases. We report performance losses (percentages) on clauses containing a connective phrase, when discourse connective phrases were excluded from clause representation building.}
\label{impact of connective phrases}
\end{table*}

\begin{table*}[]
\centering
\begin{tabular}{|l|l|ccccccc|}
\hline
\multicolumn{2}{|c|}{\multirow{2}{*}{SE Type}} & \multicolumn{7}{c|}{Predicted}                 \\ \cline{3-9} 
\multicolumn{2}{|c|}{}                         & STA   & EVE  & REP  & GENI & GENA & QUES & IMP \\ \hline
\multirow{7}{*}{Gold}          & STA           & 12558 & 980  & 32   & 931  & 155  & 51   & 85  \\
                               & EVE           & 819   & 6626 & 116  & 242  & 124  & 11   & 16  \\
                               & REP           & 42    & 143  & 1097 & 3    & 4    & 1    & 2   \\
                               & GENI          & 1157  & 175  & 3    & 4523 & 117  & 14   & 14  \\
                               & GENA          & 281   & 254  & 5    & 161  & 431  & 5    & 12  \\
                               & QUES          & 51    & 7    & 2    & 8    & 1    & 773  & 4   \\
                               & IMP           & 106   & 21   & 7    & 18   & 3    & 3    & 650 \\ \hline
\end{tabular}
\caption{Confusion Matrix of the Paragraph-level Model + CRF on the Training Set of MASC+Wiki with 10-Fold Cross-Validation.}
\label{error analysis}
\end{table*}

\subsection{Confusion Matrix}
Table \ref{error analysis} reports the confusion matrix of the full model on the training set of MASC+Wiki with cross-validation.
We can see that the four situation entity types, including two eventualities (STATE and EVENT) and two general statives (GENERIC and GENERALIZING), are often mutually confused with each other.  
To further improve the performance of situation entity type classification, it is important to accurately detect events within a clause (for fixing STATE/EVENT errors) and identify the genericity of main referents 
(for fixing STATE/GENERIC and GENERIC/GENERALIZING errors), which can be potentially achieved by incorporating linguistic features into neural net models.

\section{Conclusion}
We presented a paragraph-level neural network model for situation entity (SE) type classification which builds context-aware clause representations by modeling inter-dependencies of clauses in a paragraph. 
Evaluation shows that the paragraph-level model outperforms previous systems for SE type classification and approaches human-level performance.
In the future, we plan to incorporate SE type information in various downstream applications, e.g., many information extraction applications that require distinguishing specific fact descriptions from generic statements.

\section*{Acknowledgments}
This work was partially supported by the National
Science Foundation via NSF Award IIS-1755943.
Disclaimer: the views and conclusions contained
herein are those of the authors and should not be
interpreted as necessarily representing the official
policies or endorsements, either expressed or implied,
of NSF or the U.S. Government.
In addition, we gratefully acknowledge the support of NVIDIA Corporation for their donation of one GeForce GTX TITAN X GPU used for this research.

\bibliography{emnlp2018}
\bibliographystyle{acl_natbib_nourl}

\appendix

\end{document}